\documentclass{tlp}
\usepackage{aopmath}
\usepackage{graphicx}

\newtheorem{example}{Example} 

\begin{document}
\bibliographystyle{acmtrans}

\long\def\comment#1{}

\title[ASP and Second Language Acquisition]
{An Application of Answer Set Programming to the Field of Second Language Acquisition}

\author[D. Inclezan]
{Daniela Inclezan \\
Department of Computer Science and Software Engineering\\ 
Miami University\\
Oxford OH 45056, USA\\
E-mail: inclezd@MiamiOH.edu
}

%
%
\newcommand{\lpor}{\;\,\mbox{or}\;\,}
\newcommand{\st}{\smallskip\noindent}
\newcommand{\nmor}{\mbox{ or }}
\newcommand{\no}{\mbox{not }}

\pagerange{\pageref{firstpage}--\pageref{lastpage}}
\volume{\textbf{10} (3):}
\jdate{March 2002}
\setcounter{page}{1}
\pubyear{2002}

\submitted{1 October 2013}
\revised{19 November 2013}
\accepted{4 December 2013}

\maketitle

\label{firstpage}

\begin{abstract}
This paper explores the contributions of Answer Set Programming (ASP) 
to the study of an established theory from the field of Second Language Acquisition: 
{\em Input Processing}. The theory describes default strategies that learners 
of a second language use in extracting meaning out of a text, based on their knowledge 
of the second language and their background knowledge about the world. 
We formalized this theory in ASP, and as a result we were able to determine 
opportunities for refining its natural language description, as well as directions 
for future theory development. We applied our model to automating the prediction 
of how learners of English would interpret sentences containing the passive voice. 
We present a system, $PIas$, that uses these predictions to assist language instructors 
in designing teaching materials. 
To appear in {\em Theory and Practice of Logic Programming} ({\em TPLP}).
\end{abstract}

\section{Introduction}
\label{intro}

This paper extends a relatively new line of research that
explores the contributions of Answer Set Programming (ASP) \cite{gl91,nie98,mt99} 
to the study and refinement of {\em qualitative} scientific theories \cite{bg10,bg11}.
As pointed out by Balduccini and Girotto \citeyear{bg10},
qualitative theories tend to be formulated in natural language, often in the form of defaults.
Modeling these theories in a precise mathematical language can assist scientists 
in analyzing their theories, or
in designing experiments for testing their predictions. 
It was shown that ASP is a suitable tool for this task \cite{bg10,bg11}, 
as it provides means for an elegant and accurate 
representation of defaults, dynamic domains, and incomplete information, among others.
In our work, we explore the applicability of ASP to the formalization and analysis of
a theory from the field of Second Language Acquisition --- a discipline 
that studies the processes by which people learn a second language.\footnote{
In the field of Second Language Acquisition, the expression ``second language'' denotes 
any language that is acquired after the first one.}

{\em Our main goal is to illustrate different ways in which 
modeling the selected theory in ASP
can benefit the future development of this theory.
In particular, we will focus on contributions to 
(1) the refinement of this theory; 
(2) the automated testing of its statements; and 
(3) the development of practical applications for language teaching and testing.}

The theory we consider is VanPatten's {\em Input Processing} theory \cite{vp84,vp04}.
We chose it because it is an established theory in the field of Second Language Acquisition,
with important consequences on foreign language education.
It is specified in English in the form of a compact set of principles.
{\em Input Processing} (IP) describes the default strategies that second language learners use 
to get meaning out of text written or spoken
in the second language, during tasks focused on comprehension, 
given the learners' limitations in vocabulary, working memory, 
or internalized knowledge of grammatical structures. 
As a result of applying these strategies, even learners with limited grammatical
expertise can often, but not always, interpret input sentences correctly.
Once grammatical information is internalized, the default
strategies are overridden by the always reliable grammatical knowledge.
Hence, it can be said that IP describes an example of nonmonotonic reasoning.

IP predicts that beginner learners of English reading the sentence
{\em ``The cat was bitten by the dog''} would only be able to 
retrieve the meanings of the words {\em ``cat''}, {\em ``bitten''}, and {\em ``dog''}
and end up with something like the sequence of concepts CAT-BITE-DOG.
Although they may notice the word {\em ``was''} or the ending {\em ``-en''} 
of the verb {\em ``bitten''},
they would not be able to process them (i.e., connect them with the function
they serve, which is to indicate passive voice) 
because of limitations in processing resources.
In this context, the expression {\em processing resources} (or simply {\em resources})
refers to the amount of information that a learner can {\em hold} and {\em process}
in his/her working memory during real time comprehension of input sentences.
Additionally, IP predicts that the sentence above, now mapped into the sequence of 
concepts CAT-BITE-DOG, would be incorrectly interpreted by
these learners as {\em ``The cat bit the dog''} because of a hypothesized
strategy of assigning agent status to the first noun of a sentence.

IP, as described by VanPatten \citeyear{vp04}, consists of two principles formulated as defaults.
Each principle contains sub-principles that represent refinements of, or exceptions 
to, the original defaults. For example, a sub-principle of 
IP predicts that beginner learners of English 
would correctly interpret the sentence {\em ``The shoe was bitten by the dog''} 
because agent status cannot be assigned to the first noun, as a shoe cannot bite.
This can happen even if the learner has not yet internalized
the structure of the passive voice in English or did not have the resources
to process it in the above sentence.
Similarly for the sentence {\em ``The man was bitten by the dog''} because
it is unlikely for a man to bite a dog.
These strategies can also be applied to stories consisting of several sentences
where information from previous sentences conditions the interpretation of latter ones. 
For example, the second sentence of the story: 
{\em ``The cat killed the dog. Then, the dog was pushed by the cat.''} would
be interpreted correctly even by beginner learners, because a dead dog cannot push.
IP was shown to be applicable to other grammatical forms (e.g., clitic pronouns, subjunctive)
and other languages (e.g., Spanish, Italian, German, Chinese), 
independently from the learners' native language \cite{vp84}.

ASP is a natural choice for modeling the IP theory, first of all because
defaults and their exceptions can be represented in ASP
in an elegant and precise manner. Moreover, IP takes into consideration
the learners' knowledge about the dynamics of the world (e.g.,
people know under what conditions a biting action can occur);
in ASP, there is substantial research on how to represent actions
and dynamic domains in which change is caused by actions \cite{gl98,bg03}.
All these features of ASP were useful in creating a formalization of IP, as shown in
Section \ref{nl_analysis}. We demonstrate how the process of modeling
IP in ASP allowed us to analyze the theory's natural language description. As a result,
we were able to notice some areas that need more clarification or
could be further investigated. 
Next, we used our formalization of IP in making automated 
predictions about how learners would interpret simple sentences and paragraphs 
containing the passive voice in English. This contribution, 
described in Section \ref{pred},
can facilitate the testing of the statements of IP
or the tuning of its parameters.
Based on these predictions, we created a system, $PIas$, that can assist 
language teachers in designing instructional materials, as discussed in
Section \ref{pias}.
$PIas$ relies on the guidelines of an established teaching
method---Processing Instruction \cite{vp93,vp02}---that 
is based on the principles of {\em Input Processing}.
We end the paper with conclusions and directions for future work.

The current article extends a previous version of our work \cite{di12}.
In the remainder of the paper, we assume the reader's familiarity with ASP.

\section{An Analysis of IP Based on Its ASP Model}
\label{nl_analysis}

In this section, we describe our formalization of IP and demonstrate that
using the precise language of ASP for this purpose can highlight opportunities
for a future refinement and improvement of this theory.

\subsection{Logic Form Encoding of a Text}
\label{lp_text}

The IP theory assumes that a learner is given 
a text (called {\em input} in the enunciation of IP) --- a paragraph with one or more sentences. 
Our logic form encoding of a text uses three sorts, $words$, $sentences$, and $paragraphs$,
and two relations:

\begin{itemize}
\item $word\_of\_sent(K, S, W)$ -- the $K^{th}$ word of sentence $S$ is $W$;
\item $sent\_of\_par(K, P, S)$ -- the $K^{th}$ sentence of paragraph $P$ is $S$.
\end{itemize}

\noindent
For example, the paragraph
{\em ``The cat killed the dog. Then, the dog was pushed by the cat.''} 
in the introduction is encoded as:

$
\begin{array}{l}
sent\_of\_par(1, p, s_1). \\
sent\_of\_par(2, p, s_2). \\
word\_of\_sent(1, s_1, ``the"). \ \ \ \dots \ \ word\_of\_sent(5, s_1, ``dog").\\ 
word\_of\_sent(1, s_2, ``then"). \ \ \dots \ \ word\_of\_sent(8, s_2, ``cat").
\end{array}
$

\subsection{The First Principle of IP}
\label{p1}

Principle 1 of IP describes how likely it is for words in a sentence
to get processed by a learner engaged in a real time comprehension task,
depending on the grammatical category to which words belong.
In other words, given a sentence and a learner's knowledge of the second language,
Principle 1 predicts a possibly partial mapping of words of this sentence into cognitive concepts. 

Principle 1 makes reference to certain linguistic terms: 
a lexical item is the basic unit of the mental vocabulary
(e.g., {\em ``cat''}, {\em ``look for''}).
Content words are those that carry the meaning of a sentence: 
nouns, verbs, adjectives, and adverbs.
Forms, also called ``grammatical structures'',
are inflections, articles, or particles (e.g., 
the third-person-singular marker {\em ``-s''} 
attached to verbs as in {\em ``makes''}; the article {\em ``the''}). 
It is assumed that learners that have internalized 
a form fully also know, implicitly, whether that form is meaningful, which means that
it contributes meaning to the overall comprehension
of a sentence, or not.\footnote{An example of a nonmeaningful form is 
grammatical gender inflection in Romance languages, which manifests itself on words 
associated with nouns, such as determiners.
For instance, in Spanish, ``the moon'' is feminine (``{\em la} luna''),
while ``the sun'' is masculine (``{\em el} sol'').
The form is not meaningful because it does not reflect a ``biological difference
in the real world'' \cite{vp02b}, i.e., grammatical gender does not equal biological gender.}
Similarly, they are able to distinguish
between redundant and nonredundant meaningful forms, where a redundant form is one
whose meaning can {\em usually} be retrieved from other parts of a sentence.
Finally, the expression ``processing resources'' refers to resources 
available in the learner's working memory for holding and processing
incoming information. 

Principle 1 is formulated by VanPatten \citeyear{vp04} as follows:

\st
\textit{1. The Primacy of Meaning Principle:} {\em Learners process input 
for meaning before they process it for form.}
\begin{enumerate}
\item[\textit{1a.}] \textit{The Primacy of Content Words Principle:} {\em Learners process content
words in the input before anything else.}

\item[\textit{1b.}] \textit{The Lexical Preference Principle:} {\em Learners will tend to rely on lexical items as opposed to grammatical form to get meaning when both encode
the same semantic information.}

\item[\textit{1c.}] \textit{The Preference for Nonredundancy Principle:} {\em Learners 
are more likely to process nonredundant meaningful grammatical forms 
before they process redundant meaningful forms.}

\item[\textit{1d.}] \textit{The Meaning-Before-Nonmeaning Principle:} {\em Learners are more likely
to process meaningful grammatical forms before nonmeaningful forms irrespective of 
redundancy.}

\item[\textit{1e.}] \textit{The Availability of Resources Principle:} {\em For learners to process
either redundant meaningful grammatical forms or nonmeaningful forms, the processing
of overall sentential meaning must not drain available processing resources.}

\item[\textit{1f.}] \textit{The Sentence Location Principle:} {\em Learners tend to process items in 
sentence initial position before those in final position and these latter in turn 
before those in medial position (all other processing issues being equal).}
\end{enumerate}

\begin{example}
\label{ex1}
Let us show what predictions Principle 1 makes about the processing of words from 
the sentence:

{\em $S_1$. The cat was bitten by the dog.}

\noindent
According to 1a, content words 
have the highest chance of getting processed, in this case: 
{\em ``cat''}, {\em ``bitten''}, and {\em ``dog''}. Among them, based on 1f, 
{\em ``cat''} has the highest chance as it is in sentence initial position,
followed by {\em ``dog''} in final position, and then by {\em ``bitten''} in medial
position.

The next chance belongs to meaningful forms, based on 1d, in this case:
{\em ``the''}, {\em ``was''}, 
as well as {\em ``cat''} as an indicator of third-person singular, 
and {\em ``bitten''} (more precisely the suffix {\em ``-en''}) 
as an indicator of passive voice.
According to 1c, out of these forms, the nonredundant ones are more likely to get processed,
in particular the definite article {\em ``the''} 
and the word {\em ``cat''} as an indicator of third-person singular,
both in initial position, followed by {\em ``the''} in final sentence position,
and then by the forms {\em ``was''} as an indicator of past tense and {\em ``bitten''} 
as an indicator of passive voice in medial position.

Principle 1e says that the whole sentence has the next chance of getting processed, followed
by the redundant form {\em ``by''}. Finally, according to 1b, 
the redundant form {\em ``was''} (i.e., the suffix {\em ``-s''})
as an indicator of third-person singular may or may not
get processed, independently of available resources, because its meaning
was already obtained from the word {\em ``cat''}.
Note that, how many words actually get processed 
depends on the resource capacity of a learner.
\end{example}

\st
{\bf Encoding a Learner's Knowledge of the Second Language}

\noindent
The IP theory is supposed to be applicable independently from the
mental model of a second language that is assumed \cite{vp04}.
This allows us to make the simplification of {\em not} considering inflections 
on a word (e.g., {\em ``-s''}, {\em ``-en''}) separately from the rest of the word. 
As a result, a word can be viewed as belonging to multiple categories.
For instance, {\em ``makes''} can be viewed as a content word 
referring to the action of making something; it can also be perceived as
a form indicating that the doer is not the speaker nor the addressee
and that the action is occurring in the present (due to the ending {\em ``-s''}).

Given the categories listed in Principle 1, we divide
$words$ into two subclasses, $content\_words$ and $forms$;
$forms$ are divided into $m\_forms$ (meaningful) and $nm\_forms$ (nonmeaningful),
while $m\_forms$ are further divided into $r\_m\_forms$ (redundant) and $nr\_m\_forms$
(nonredundant). The leaves of this hierarchy are denoted by a special sort,
$leaf\_ctg$. All nodes of the hierarchy are denoted by the sort $category$.
We introduce a new sort, $concept$, denoting 
language-independent cognitive concepts, 
such as entities, actions, or semantic concepts (e.g., the concepts of past tense
and passive voice). 
Additionally, we specify a learner's knowledge of the second language using:

\begin{itemize}
\item $in(W, Ctg)$ -- word $W$ belongs to category $Ctg$;
\item $meaning(W, Ctg, C)$  -- word $W$ interpreted as a member of category $Ctg$ has the meaning $C$.
\end{itemize}

\medskip

\st
{\bf Our ASP Model of Principle 1}

\noindent
After careful analysis, it is clear that Principle 1 
specifies a partial order between words
in a sentence, given the category they belong to and their sentential position.
Greater elements in this ordering have more 
chances of being processed than lesser elements.
{\em It is important to note that we only realized that
a partial order was described when attempting to formulate Principle 1 in ASP.
The fact was not immediately obvious to us because 
the sub-principles specifying an order based on word categories
(1a, 1c, 1d) and sentential position (1f) are not grouped together
in the text of the theory.\footnote{
Sub-principles 1b and 1e describe constraints that
can further limit the chances of a word to get processed.}
Hence, we can say that modeling IP in ASP led us 
to a better understanding of the theory.}

We start formalizing Principle 1 by looking at its sub-principles
1a, 1c, and 1d, which describe a partial order on word categories.
To model it, we define a relation $is\_ml\_ctg$
on categories, where $is\_ml\_ctg(Ctg_1, Ctg_2)$ says that words from category 
$Ctg_1$ are more likely to get processed than words from category $Ctg_2$.
Based on Principles 1a, 1d, and 1c, respectively, we have the facts:
$$
\begin{array}{l}
is\_ml\_ctg(content\_words, forms).\\
is\_ml\_ctg(m\_forms, nm\_forms).\\
is\_ml\_ctg(nr\_m\_forms, r\_m\_forms).
\end{array}
$$
Next, we look at Principle 1f, which describes a similar partial order on
sentence positions. To specify the different possible sentence positions, 
we define a sort $sentence\_position$ with three elements: $initial$, $medial$,
and $final$. We use a relation $is\_ml\_pos(Pos_1, Pos_2)$,
which says that words in sentence position $Pos_1$
are more likely to be processed than words in $Pos_2$ (as long as they
belong to the same word category). We encode Principle 1f via the facts:
$$
\begin{array}{c}
is\_ml\_pos(initial, final).\\
is\_ml\_pos(final, medial).
\end{array}
$$
By $ml\_ctg$ and $ml\_pos$, respectively, we denote the transitive closures 
of the two relations above. In addition, we extend the relation $ml\_ctg$ down 
to subclasses of categories, but not upwards to superclasses.

Based on the two relations above, we can now define the partial relation 
between words, given their category and position. 
{\em Our modeling process illuminated the fact that the IP theory 
does not say how many words starting from the beginning
of a sentence are part of the ``sentence initial position.'' 
This expression needs to be precisely defined in the future.}
For the moment, we define initial position as the first $n$ words of a sentence, 
where $n$ is a parameter of the encoding. 
Similarly for final positions. 
We use the relation $pos(K, S, Pos)$ to say that the $K^{th}$
word of sentence $S$ is in $Pos$ sentence position.
We introduce a relation $ml\_wrd(K_1, S, Ctg_1, K_2, Ctg_2)$,
which says that the $K_1^{th}$ word
of $S$ is more likely to get processed for its interpretation as an element of
the category $Ctg_1$ than the $K_2^{th}$ word of the same sentence
for category $Ctg_2$:
$$
\begin{array}{lll}
ml\_wrd(K_1, S, Ctg_1, K_2, Ctg_2) & \leftarrow & leaf\_ctg(Ctg_1),\ \ leaf\_ctg(Ctg_2),\\
	                                 &            & word\_of\_sent(K_1, S, W_1),\ \ in(W_1, Ctg_1),\\
                                   &            & word\_of\_sent(K_2, S, W_2),\ \ in(W_2, Ctg_2),\\
                                   &            & ml\_ctg(Ctg_1, Ctg_2).\\
ml\_wrd(K_1, S, Ctg, K_2, Ctg) & \leftarrow & leaf\_ctg(Ctg),\\
                               &            & word\_of\_sent(K_1, S, W_1),\ \ in(W_1, Ctg),\\
                               &            & word\_of\_sent(K_2, S, W_2),\ \ in(W_2, Ctg),	\\
                               &            & pos(K_1, S, Pos_1),\ \ pos(K_2, S, Pos_2),\\
                               &            & ml\_pos(Pos_1, Pos_2).
\end{array}
$$
The first rule relates to Principles 1a, 1c, and 1d, as it is based on the ordering of
categories; the second rule is about Principle 1f, as it uses the sentence position
ordering, for a given category of words. 
The effects of the ordering $ml\_wrd$ on the processing of words of 
a sentence will be seen later. 

Next, we specify that, normally, a word will get processed
(i.e., be mapped into a concept) if enough resources are available. We introduce a relation 
$map(K$, $S$, $Ctg, C)$, which says that the 
$K^{th}$ word of $S$ was processed according to category $Ctg$ and was mapped into concept
$C$. We encode Principle 1 as:
\begin{equation}
\begin{array}{lll}
map(K, S, Ctg, C) & \leftarrow & word\_of\_sent(K, S, W), \ in(W, Ctg),\\
                  &            & leaf\_ctg(Ctg), \ meaning(W, Ctg, C),\\
                  &            & enough\_resources\_available(K, S, Ctg),\\
                  &            & \no ab(d_{map}(K, S, Ctg, C)).
\end{array}
\label{eq_nondet1}
\end{equation}
{\em The IP theory does not give any details about the initial resources in working memory
available to a learner for processing a sentence, nor about how
learners at different levels of proficiency consume 
those resources while attaching meaning to words. This
is another aspect of the theory that needs more careful consideration.}
To solve this issue, we created a simple model of resources, 
in which we assume a fixed resource capacity available per sentence;
this capacity decreases by one unit with each association of meaning to a word.
We introduced a predicate $resources\_consumed(N, K, S, Ctg)$, which says that
$N$ resources are consumed in processing those words that are more likely to get processed
than the $K^{th}$ word of sentence $S$ for category $Ctg$. 
The definition of this relation captures the implications 
of the ordering $ml\_wrd$ on the processing of words:
$$
\begin{array}{l}
resources\_consumed(N, K, S, Ctg) \leftarrow \\
\ \ \ \ \ \ \ \ \ \ \ \ \ \ \ \ word\_of\_sent(K, S, W), \ \ in(W, Ctg), \ \ leaf\_ctg(Ctg),\\
\ \ \ \ \ \ \ \ \ \ \ \ \ \ \ \ N = \#count\ \{ ml\_wrd(K_1, S, Ctg_1, K, Ctg) :\\
\ \ \ \ \ \ \ \ \ \ \ \ \ \ \ \ \ \ \ \ \ \ \ \ \ \ \ \ \ \ \ \ \ \ \ leaf\_ctg(Ctg1)\ \}. 
\end{array}
$$
Although this model does not reflect the complexities of working memory,
it is enough for our purposes, as the IP theory only focuses on 
the expected {\em end result} of processing a sentence 
in working memory: certain word-to-concept associations will be made while others will not.
To model Principle 1e, we assume that processing the whole sentence
decreases available resources by one unit.

The only remaining principle is 1b. Based on its accompanying explanation
provided by VanPatten \citeyear{vp04}, its meaning is that a form that is {\em normally} redundant may not be processed
at all if it is {\em actually} redundant in that sentence (i.e.,
its meaning was already extracted from some other word).
We encode this knowledge as a {\em possible} weak exception to the default in the rule
for predicate $map$, via a disjunctive rule:
\begin{equation}
\begin{array}{l}
ab(d_{map}(K, S, Ctg, C)) \ \nmor \ \neg ab(d_{map}(K, S, Ctg, C)) \leftarrow \\
\ \ \ \ \ \ \ \ \ \ \ \ \ \ \ \ \ \ word\_of\_sent(K, S, W),\\
\ \ \ \ \ \ \ \ \ \ \ \ \ \ \ \ \ \ meaning(W, Ctg, C),\\
\ \ \ \ \ \ \ \ \ \ \ \ \ \ \ \ \ \ word\_of\_sent(K_1, S, W_1),\\ 
\ \ \ \ \ \ \ \ \ \ \ \ \ \ \ \ \ \ K \neq K_1,\ \ W \neq W_1,\\
\ \ \ \ \ \ \ \ \ \ \ \ \ \ \ \ \ \ map(K_1, S, Ctg_1, C).
\end{array}
\label{eq_nondet2}
\end{equation}
The informal reading of this axiom is that meanings that are actually
redundant in a sentence may or may not be exceptions to the default for relation $map$.

\subsection{The Second Principle of Input Processing}
\label{p2}

Principle 2 describes the strategies that learners employ to understand the 
meaning of a sentence. 
The input of Principle 2 is the output of Principle 1 for a given sentence
(i.e., a mapping of words to concepts), together with the learner's background 
knowledge about the world. Its output is an event
denoting the meaning extracted by the learner from that sentence.
When considering a story consisting of several sentences, 
the output of Principle 2 is a series of events that correspond to the sentences
in that paragraph. For simplicity, we assume here that each sentence
describes a single event, and that sentences of a story describe events in the
order in which those events happened. 
Principle 2 is formulated by VanPatten \citeyear{vp04,vp02} as follows:

\st
{\em 2. The First Noun Principle (FNP): Learners tend to process the first
noun or pronoun they encounter in a sentence as the agent.}
\begin{enumerate}
\item[{\em 2a.}] {\em The Lexical Semantics Principle: Learners may rely on lexical semantics,\footnote{
{\em Lexical semantics} refers to the meaning of lexical items.}
where possible, instead of on word order to interpret sentences.}

\item[{\em 2b.}] {\em The Event Probabilities Principle: Learners may rely on event
probabilities, where possible, instead of on word order to interpret sentences.}

\item[{\em 2c.}] {\em The Contextual Constraint Principle: Learners may rely less on the
First Noun Principle if preceding context constrains the possible
interpretation of a clause or sentence.}

\item[{\em 2d.}] {\em Prior Knowledge: Learners may rely on prior knowledge,
where possible, to interpret sentences.}

\item[{\em 2e.}] {\em Grammatical Cues: Learners will adopt other processing strategies for 
grammatical role assignment only after their developing system\footnote{
{\em Developing system} refers to the representation of
grammatical knowledge in the mind of the second language learner. 
This representation changes as 
the learner acquires more knowledge.} 
has incorporated other cues.}
\end{enumerate}

\begin{example}
\label{ex2}
We illustrate the predictions made by Principle 2 for several sentences.
First, we consider the case of beginner learners, who have limited
resources and vocabulary, and can only process the content words out of a sentence.
Based on Principle 1, beginners would map the words 
{\em ``cat''}, {\em ``bitten''}, and {\em ``dog''}
in sentence $S_1$ from Example \ref{ex1} into the concepts 
CAT, BITE, and DOG
respectively
and would not be able to process any other words.
Principle 2 predicts that beginners would assign agent status to the first
noun in $S_1$ and hence interpret $S_1$ incorrectly as
{\em ``The cat bit the dog.''}

Beginners are expected to correctly interpret the sentence:

{\em $S_2$. The shoe was bitten by the dog.}

\noindent
as a shoe cannot bite a dog (lexical semantics). Based on Principle 2a,
lexical semantics override the assignment of agent status to the first noun.
The sentence:

{\em $S_3$. The man was bitten by the dog.}

\noindent
is also supposed to be interpreted correctly by beginners because men
normally do not bite animals (event probabilities and Principle 2b).

Principle 2d predicts the correct interpretation of:

{\em $S_4$. Holyfield was bitten by Tyson.}

\noindent
assuming that learners have the prior knowledge that Tyson bit Holyfield. 

Let us now consider some short paragraphs:

$
\begin{array}{ll}
P_1\mbox{.} & (S_5\mbox{.}) \mbox{\em{ The cat pushed the dog. }} 
(S_6\mbox{.}) \mbox{\em{ Then, the dog 
was bitten by the cat.}}
\end{array}
$

\noindent
Sentence $S_6$ is supposed to be incorrectly interpreted by beginners because none of the
Principles 2a-e applies.
Instead, the second sentence of the paragraph:

$
\begin{array}{ll}
P_2\mbox{.} & (S_7\mbox{.}) \mbox{\em{ The cat killed the dog. }} 
(S_8\mbox{.}) \mbox{\em{ Then, the dog was pushed by the cat.}}
\end{array}
$

\noindent
would be interpreted correctly due to lexical semantics in context, as predicted
by Principles 2a and 2c together.

Finally, let us consider advanced learners who possess enough resources and a large vocabulary,
which allow them to map all words of a sentence into concepts. 
According to Principle 2e, these learners are expected to interpret all above sentences correctly, 
as they are able to detect the use of the active or passive voice and
can rely on grammatical cues for sentence interpretation. 
\end{example}

\st
{\bf Encoding a Learner's Background Knowledge about the World}

\noindent
Learners are assumed to possess some background knowledge about the world and its
dynamics. Three important types of information are supposed to be derivable from
this knowledge base, and we capture them using the predicates: 

\begin{itemize}
\item $impossible(Ev, I)$ -- event $Ev$ is physically impossible to occur 
at step $I$ of the narrated story;
\item $unlikely(Ev, I)$ -- event $Ev$ is unlikely to occur at step $I$ of the narrated story;
\item $hpd(Ev)$  -- event $Ev$ is known to have happened in reality.
\end{itemize}

\noindent
To model the background knowledge base of a learner, we use 
known methodologies for representing dynamic domains in ASP \cite{gl98,bg03}.
As a result, atoms of the type $impossible(Ev, I)$ are derived from
axioms specifying preconditions for the execution of actions (i.e., 
{\em executability conditions}); $unlikely(Ev, I)$ atoms are obtained from axioms
encoding default statements and their exceptions \cite{bg94};
$hpd(Ev)$ atoms are simply stored as a collection of facts.\footnote{We  
considered using probabilistic ASP languages such as P-log \cite{cgr09} to 
model event probabilities. However, we decided against it because we believe that
our naive model is closer to how humans record information in their minds,
and because finding the exact probability of an event 
(e.g., how likely it is for a cat to bite a dog) is a complex task in itself.}
Note that our formalization of the IP theory is independent from the 
underlying model of the world and its dynamics. This means that other models
could be easily coupled with our formalization of Principle 2, as long as the model
can derive atoms of the three types mentioned above.

\medskip\noindent
{\bf Our ASP Model of Principle 2}

\noindent
We assume that each sentence in the input describes exactly one 
event, and that the $N^{th}$ sentence of a paragraph describes the $N^{th}$ occurring event.

We start by introducing some terminology. 
By the {\em direct (reverse) meaning} of a sentence we mean the action 
denoted by the verb of the sentence, and whose agent is the entity denoted by 
the {\em first} ({\em second}) noun appearing in the
sentence.
For instance the direct meaning of {\em ``The dog was bitten by the cat''} is
the event of ``the dog biting the cat,'' while its reverse meaning is the event
of ``the cat biting the dog.''
We use the predicate $dir\_rev\_m(Dir, Rev, S)$ 
to say that $Dir$ is the direct meaning and $Rev$ is the reverse meaning of sentence $S$.

Principle 2, also called the First Noun Principle (FNP),
is a default statement and its sub-principles express exceptions to it.
To encode Principle 2, we use a relation $extr\_m(Ev, S, fnp)$
saying that the learner extracted the meaning $Ev$ from sentence $S$
by applying FNP:
$$
\begin{array}{lll}
extr\_m(Dir, S, fnp) & \leftarrow & \no extr\_m(Rev, S, fnp),\\
			               &            & dir\_rev\_m(Dir, Rev, S).
\end{array}
$$
The rule says that learners applying the FNP
will extract the direct meaning from a sentence,
unless they extract the reverse meaning.

We represent Principle 2a using the axiom:
$$
\begin{array}{lll}
extr\_m(Rev, S, fnp) & \leftarrow & impossible(Dir, I),\\
 & & \no impossible(Rev, I),\\
 & & dir\_rev\_m(Dir, Rev, S),\\
 & & sent\_of\_par(I, P, S).
\end{array}
$$
Informally, it says that learners will assign the reverse meaning to
a sentence if this is a possible meaning, and the direct meaning 
is impossible.

The formalization of Principle 2b will be similar:
$$
\begin{array}{lll}
extr\_m(Rev, S, fnp) & \leftarrow & \no impossible(Dir, I),\\
 & & unlikely(Dir, I),\\
 & & \no hpd(Dir),\\
 & & \no impossible(Rev, I),\\
 & & \no unlikely(Rev, I),\\
 & & dir\_rev\_m(Dir, Rev, S),\\
 & & sent\_of\_par(I, P, S).
\end{array}
$$
I.e., a sentence will be assigned its reverse meaning
if the direct meaning is possible, but unlikely and not known to have actually happened,
while the reverse meaning may hypothetically occur (i.e., it is possible and not unlikely).

Principle 2d is encoded as follows:
$$
\begin{array}{lll}
extr\_m(Rev, S, fnp) & \leftarrow & hpd(Rev),\\ 
                  &            & dir\_rev\_m(Dir, Rev, S),\\
                  & &  sent\_of\_par(I, P, S).
\end{array}
$$
This says that a learner using the FNP will
extract the reverse meaning if he knows that this event
actually happened.

The preference for grammatical cues when such cues can be interpreted
(Principle 2e) is encoded via the rules:
$$
\begin{array}{lll}
extr\_m(Ev, S) & \leftarrow & extr\_m(Ev, S, grm\_cues).\\
extr\_m(Ev, S) & \leftarrow & extr\_m(Ev, S, fnp),\\
               &            & \no extr\_m\_by(S, grm\_cues).\\
extr\_m\_by(S, X) & \leftarrow & extr\_m(Ev, S, X).
\end{array}
$$
where $extr\_m(Ev, S)$ says that $Ev$ is the meaning extracted from $S$; 
$extr\_m(Ev$, $S$, $grm\_cues)$ -- the meaning $Ev$ was
extracted from $S$ based on grammatical cues (which vary for different
grammatical forms); and $extr\_m\_by(S, X)$ --
the meaning of $S$ was extracted based on strategy $X$. 
The definition of $extr\_m(Ev, S$, $grm\_cues)$, not shown here,
captures the fact that different grammatical forms have different grammatical cues.
For instance, the cues for passive voice in English are the past participle (e.g., {\em ``bitten''})
and the passive voice auxiliary (e.g., {\em ``was''}).
An $extr\_m(Ev, S, grm\_cues)$ atom belongs to an answer set if the learner was able to
map the main grammatical form(s) in the sentence (in the case of passive voice, 
the past participle and the passive voice auxiliary) into the corresponding 
abstract concept (here, passive voice), 
and if $Ev$ is the correct interpretation of sentence $S$.

In our formalization of FNP, Principle 2c was embedded in the representation of Principles
2a, 2b, and 2d. The one thing left for contextual constraints is
to record the events corresponding to the meaning extracted from previous sentences of the story,
assuming the first time step of the story is 1. 
$$
\begin{array}{lll}
occurs(Ev, I) & \leftarrow & extr\_m(Ev, S),\\
              &            & sent\_of\_par(I, P, S).
\end{array}
$$
{\em Notice that Principle 2c specifies that preceding sentences in a paragraph
constrain the interpretation of latter sentences, but does not mention
a possible effect of {\em succeeding} sentences on the re-interpretation
of {\em earlier} sentences that were initially processed incorrectly.
Also, Principle 2 in general does not address sentences that describe events which cannot
physically take place in the real world, unless understood metaphorically (e.g.,
{\em ``The dog was bitten by the shoe.''}).
These are interesting directions of research that the IP theory could address.}

\section{Automating the Predictions of IP} 
\label{pred}

We used our model of the IP theory to generate automated predictions about 
how sentences like the ones in Examples \ref{ex1} and \ref{ex2}
would be interpreted by learners of English.
We considered two different types of learners: advanced and beginners.
They both shared the same background knowledge about the world, 
but had different knowledge about the second language.
The advanced learner internalized the meaning of all content words and forms
in our vocabulary, whereas beginners would only master the meaning of content words,
but not forms.
For each type of learner, we created a logic program $\Pi$ (indexed by either
$adv$ or $beg$) by putting together 
the corresponding knowledge of the second language,
the background knowledge about the world, 
and the formalizations of the two principles. 
For any text X, by $lp(X)$ we denote the logic form encoding of $X$ as
presented in Section \ref{lp_text}.
{\em The answer set(s) of the program $\Pi \cup lp(X)$
corresponds to predictions of the IP theory about how a learner would interpret $X$.}

We first run tests for Principle 1 by using 
sentence $S_1$, 
copied here with its words 
annotated by their sentential indices for a better understanding of the results:
$$
S_1\mbox{.}\ The_1\ cat_2\ was_3\ bitten_4\ by_5\ the_6\ dog_7\mbox{.}
$$
We set the sentence position parameter $n$ to 2,
and run experiments for different resource capacities.
{\em A scientist working on a refinement of IP theory
could easily change the values of these parameters and thus use
our formalization of IP to fine-tune them.}

The answer sets of the program $\Pi_{adv} \cup lp(S_1)$, computed using the ASP solver
{\sc claspD} \cite{drgegrkakoossc08a}, contained
the $map$ facts presented in Table 1.
An atom $map(k, s_1, ctg, c)$ in the answer set for capacity $m$ indicates
that the $k^{th}$ word of $S_1$ will get processed by an advanced learner with capacity $m$,
and be mapped into the cognitive concept $c$.
The atom $map(3,s_1,r\_m\_forms,third\_person\_singular)$ 
on the last line of the table is marked with an asterisk
because two answer sets are generated for capacity 11, 
and this atom is part of one answer set but not the other.
The non-determinism comes from the disjunctive rule (\ref{eq_nondet2}) that,
together with rule (\ref{eq_nondet1}), encodes Principle 1b
stating that learners tend to extract meaning from 
content words rather than from forms when they both encode the same meaning.
Rule (\ref{eq_nondet2}) specifies that learners may or may not obey this default.
In the case of $S_1$, the form {\em ``was''} indicates (among other things) 
that the event is about an entity other than the speaker and the addressee 
(i.e., third person singular). 
However, this meaning was already extracted by the learner from the content word {\em ``cat''}
(see atom $map(2,s_1,nr\_m\_forms,third\_person\_singular)$ for capacity 9), hence
the form {\em ``was''} may or may not be processed for the same meaning.
Another thing to note is that a beginner learner would not be able to
process grammatical forms in the sentence even if s/he had a capacity exceeding value 11,
just because s/he has not yet internalized forms.
\begin{table}
\caption{Automated Predictions for Principle 1}
\label{table1}
    \begin{minipage}{\textwidth}
    \begin{tabular}{cl}
      \hline\hline
      {\bf Capacity} & {\bf Additional $map$ Facts w.r.t. Answer Sets for Smaller Capacities}\\ 
      \hline
      0 & $\emptyset$\\
      \noalign{\vspace {.2cm}}
      1 & $map(2,s_1,content\_words,cat)$\\
      \noalign{\vspace {.2cm}}
      2 & $map(7,s_1,content\_words,dog)$\\
      \noalign{\vspace {.2cm}}
      3 & $map(4,s_1,content\_words,bite)$\\
      \noalign{\vspace {.2cm}}
      9 & $map(1,s_1,nr\_m\_forms,definite)$ \\
        & $map(2,s_1,nr\_m\_forms,third\_person\_singular)$ \\
        & $map(6,s_1,nr\_m\_forms,definite)$\\
        & $map(3,s_1,nr\_m\_forms,passive\_voice)$ \\
        & $map(3,s_1,nr\_m\_forms,past\_tense)$ \\
        & $map(4,s_1,nr\_m\_forms,past\_participle)$\\
      \noalign{\vspace {.2cm}}
      11 & $map(5,s_1,r\_m\_forms,agency)$\\
         & $map(3,s_1,r\_m\_forms,third\_person\_singular) ^\ast$\\
      \hline\hline
    \end{tabular}
    \vspace{-2\baselineskip}
  \end{minipage}
\end{table}

Next, we tested our predictions for Principle 2 for beginners and advanced learners.
In both cases, we set the resource capacity to  value 11.
The relevant parts of the answer sets for the texts in Example \ref{ex2} 
can be seen in Tables 2 and 3, where terms like $ev(bite, cat, dog)$ are used to denote
events, in this case ``a cat biting a dog''.
We do not show all the predictions for advanced learners, because they are expected to
interpret all sentences and paragraphs correctly.
\begin{table}
\caption{Automated Predictions for Principle 2 and Beginner Learners}
\label{table2}   
    \begin{minipage}{\textwidth}
    \begin{tabular}{clc}
      \hline\hline
      $\ X\ $ & {\bf Answer Set of $\Pi_{beg} \cup lp(X)$ contains} & {\bf $\ \ $Is X interpreted correctly?}\\ \hline
      $S_1$ & $extr\_m(ev(bite,cat,dog), s_1)$ & NO \\
            & $extr\_m\_by(s_1, fnp)$ &  \\
      \noalign{\vspace {.2cm}}
      $S_2$ & $extr\_m(ev(bite,dog,shoe), s_2)$ & YES \\
            & $impossible(ev(bite, shoe, dog), 1)$ &  \\
      \noalign{\vspace {.2cm}}
      $S_3$ & $extr\_m(ev(bite,dog,man), s_3)$  & YES \\
            & $unlikely(ev(bite, man, dog), 1)$  &  \\
      \noalign{\vspace {.2cm}}
      $S_4$ & $extr\_m(ev(bite,tyson, holyfield), s_4)$ & YES \\
            & $hpd(ev(bite, tyson, holyfield))$ &  \\
      \noalign{\vspace {.2cm}}
      $P_1$ & $extr\_m(ev(push, cat, dog), s_5)$  & NO \\
            & $extr\_m(ev(bite, dog, cat), s_6)$  &  \\
      \noalign{\vspace {.2cm}}
      $P_2$ & $extr\_m(ev(kill, cat, dog), s_7)$  & YES \\
            & $extr\_m(ev(push, cat, dog), s_8)$  &  \\
            & $impossible(ev(push, dog, cat), 2)$  & \\
      \hline\hline
    \end{tabular}
    \vspace{-2\baselineskip}
  \end{minipage}
\end{table}
\begin{table}
\caption{Automated Predictions for Principle 2 and Advanced Learners}
\label{table3}   
    \begin{minipage}{\textwidth}
    \begin{tabular}{clc}
      \hline\hline
      $\ X\ $ & {\bf Answer Set of $\Pi_{adv} \cup lp(X)$ contains} & {\bf $\ \ $Is X interpreted correctly?}\\ \hline
      $S_1$ & $extr\_m(ev(bite, dog, cat), s_1)$  & YES \\ 
            & $extr\_m\_by(s_1, grm\_cues)$  &  \\ 
      \hline\hline
    \end{tabular}
    \vspace{-2\baselineskip}
  \end{minipage}
\end{table}

Our automated predictions matched the ones in Examples \ref{ex1} and \ref{ex2},
which suggests that our model of IP is correct.

\section{The System $PIas$}
\label{pias}

We created a system, $PIas$, designed
to assist instructors in preparing materials for 
the passive voice in English. $PIas$ follows the guidelines of a successful teaching
method called Processing Instruction (PI) \cite{vp93,vp02,vp02b,lvp03}, 
developed based on the principles of IP. 
For a sentence to be {\em valuable} in this approach, 
it must lead to an {\em incorrect} interpretation when grammatical cues are {\em not} used
but the FNP is. In other words, learners must be made aware that their default
strategies can sometimes be counterproductive, whereas grammatical cues
are always reliable.
$S_1$ above is an example of a valuable sentence; $S_2$, $S_3$, and $S_4$ are not.

$PIas$ has two functions. The first one is to specify whether sentences and paragraphs
created by instructors are valuable or not. 
{\em This is relevant because even instructors trained in PI happen
to create bad materials.}\footnote{A non-valuable sentence crafted by a 
researcher from the Second Language Acquisition community \cite{q08} 
is {\em ``The ball was pushed by the rabbit.''}
This sentence is not valuable \cite{visf09} because event probabilities ensure that
even beginner learners will interpret this
sentence correctly -- it is more likely for a rabbit to push a ball than
for a ball to push a rabbit.}
We define: 
$$
\begin{array}{lll}
valuable(S) & \leftarrow & extr\_m(Ev_1, S, grm\_cues),\\
            &            & extr\_m(Ev_2, S, fnp),\\
            &            & Ev_1 \neq Ev_2.
\end{array}
$$
We create a module ${\cal M}$ containing this definition and its extension
to paragraphs. 
$PIas$ takes as an input a sentence or paragraph $X$
in natural language, encodes it in its logic form $lp(X)$, and computes the answer sets
of a program consisting of $\Pi_{adv}$, ${\cal M}$,
and $lp(X)$. $X$ is valuable if the atom $valuable(X)$ belongs to all   
answer sets of the resulting program.

The second function of $PIas$ is to generate all valuable sentences given 
a vocabulary and some simple grammar.
{\em This is important because PI requires to expose learners to a large number of
valuable sentences.}\footnote{In their study, VanPatten and Cadierno \citeyear{vpc93} used 120 
valuable sentences for a single grammatical form.} 
We add to ${\cal M}$ rules for sentence creation.
For instance, one particular type of sentence is generated by: 
$$
\begin{array}{l}
sentence(s(``The", N_1, ``was", V, ``by", ``the", N_2)) \leftarrow \\
\ \ \ \ \ \ \ \ \ \ \ \ \ \ \ \ \ \ \ schema(N_1, V, N_2).\\
word\_of\_sent(1, s(``The", N_1, ``was", V, ``by", ``the", N_2), ``the") \leftarrow \\
\ \ \ \ \ \ \ \ \ \ \ \ \ \ \ \ \ \ \ schema(N_1, V, N_2).
\end{array}
$$
where $schema(N_1, V, N_2)$ is true if $N_1$ and $N_2$ are common nouns
and $V$ is a verb in the past participle form (e.g., {\em ``bitten''}).
Atoms of the type $valuable(S)$ in the answer set(s) of the program $\Pi_{adv} \cup {\cal M}$
give all the valuable sentences that can be generated using our grammar and vocabulary.

The $PIas$ system is currently just a proof-of-concept, as it only
generates simple sentences, given a controlled grammar and a small vocabulary.
Its evaluation was done by the author, who was previously trained in the
Processing Instruction teaching method and was involved in research in the
field of Second Language Acquisition \cite{visf09}.
In the future, we plan to expand the system in order to make it capable of creating
more complex sentences and stories containing the passive voice,
as well as complete teaching and testing activities that interleave sentences
containing the target grammatical form with sentences that do not contain it.
Once $PIas$ is capable of producing such activities, we plan to subject
the system to a more rigorous evaluation.

\section{Conclusions and Future Work}
\label{future_work}

This paper has shown three different directions in which modeling
an important theory from the field of Second Language Acquisition 
can contribute to the development of this theory. 
First of all, we identified aspects in the text of the theory description that 
need refinement (i.e., the definition of ``sentence initial position'';
the presentation of Principle 1, 
whose sub-principles could be ordered differently to facilitate 
a deeper understanding)
and determined opportunities for future theory development
(i.e., How are resources in working memory consumed during comprehension tasks?
Do succeeding sentences in a narrative constrain 
the interpretation of previous sentences?
How are non-sense sentences interpreted?)
Second, we have shown how our ASP model can be used to design experiments
for testing this theory and fine-tuning its parameters.
Third, we described a system, $PIas$ that assesses the quality
of materials created by language instructors, and creates valuable sentences.

{\em We hope that the application presented here, and its three main contributions,
will help promote ASP as a tool for the study of qualitative 
theories, in different fields.}
To the best of our knowledge, the only other uses of ASP for the refinement
of the natural language description of a cognitive theory are the papers
of Balduccini and Girotto \citeyear{bg10,bg11} that inspired the current work.
In the field of Applied Linguistics, we are aware of the use of
computer models in simulating theory predictions \cite{dvhg98,dvh02}.
The mentioned computer models were created using procedural languages.
In contrast to these approaches, our main focus is on facilitating 
the revision of the {\em description of a theory} by formalizing it
in a mathematical language.
ASP showed to be particularly suitable for this task because
it allows for a precise and elegant encoding of defaults, uncertainty, and
evolving domains.
For us, the simulation of results and the automated testing of the theory's predictions 
is just a consequence of our primary goal.

Our principal interest in expanding the work in this paper will be on improving the capabilities of
the $PIas$ system. We want $PIas$ to use the valuable 
sentences it generates in creating complete exercises or activities suitable for teaching. 
We also intend to make $PIas$ capable of producing valuable {\em paragraphs}. 
One difficult question to address here will be {\em What makes a collection
of sentences a story?}

\subsubsection*{Acknowledgments}
The author warmly thanks Michael Gelfond, Marcello Balduccini, 
and the anonymous reviewers for their valuable suggestions.

\bibliography{ip}

\end{document}